\documentclass[11pt,a4paper]{article}
\usepackage[margin=1in]{geometry}
\usepackage[T1]{fontenc}
\usepackage{newtxtext}
\usepackage{amsmath}
\usepackage{amsthm}
\usepackage[varvw]{newtxmath}
\usepackage{graphicx,booktabs,array,tabularx,tikz,listings}
\usepackage[numbers,sort&compress]{natbib}
\usepackage{xurl}
\usepackage[hidelinks]{hyperref}
\usetikzlibrary{arrows.meta,fit,positioning,calc}
\hypersetup{pdftitle={One Monitor Does Not Fit All: Formalism-Agnostic Runtime Verification for Agentic AI},
  pdfauthor={Angelo Ferrando},
  pdfsubject={Runtime verification and enforcement for agentic AI},
  pdfkeywords={runtime verification, agentic AI, runtime enforcement, monitor interoperability}}
\theoremstyle{definition}
\newtheorem{definition}{Definition}
\theoremstyle{plain}
\newtheorem{proposition}{Proposition}
\providecommand{\doi}[1]{\href{https://doi.org/#1}{\nolinkurl{#1}}}
\newcommand{\tool}[1]{\texttt{#1}}
\newcommand{\AgenticRV}{\textsc{AgenticRV}}
\newcommand{\obs}{\mathit{obs}}
\newcommand{\eps}{\varepsilon}
\newcommand{\permit}{\ensuremath{\mathsf{permit}}}
\newcommand{\reject}{\ensuremath{\mathsf{reject}}}
\newcommand{\allow}{\ensuremath{\mathsf{allow}}}
\newcommand{\block}{\ensuremath{\mathsf{block}}}
\newcolumntype{L}[1]{>{\raggedright\arraybackslash}p{#1}}
\begin{document}
\title{One Monitor Does Not Fit All:\texorpdfstring{\\}{ }Formalism-Agnostic Runtime Verification for Agentic AI}
\author{Angelo Ferrando\\[0.3em]
\small University of Modena and Reggio Emilia, Modena, Italy\\
\small\href{mailto:angelo.ferrando@unimore.it}{\texttt{angelo.ferrando@unimore.it}}}
\date{}
\maketitle
\begin{abstract}
Agentic AI systems combine model decisions, mutable state, tool calls, and
delegation to external agents. Monitoring them requires observations from
different execution boundaries, while prevention requires control before an
operation produces its effect. We present \AgenticRV, a runtime-verification
architecture that separates these boundaries from the monitoring formalism.
Adapters emit canonical events; each monitor receives its own projection and
abstraction; and a decision policy uses normalised verdicts to select a reaction
supported by the responsible adapter. We formalise when observations suffice,
monitor replacement preserves decisions, and blocking prevents unsafe effects.
We also distinguish rejected
attempts from completed effects, avoiding the assumption that a blocked request
must leave every monitor in a permanent violation state. The Python
implementation supports LangGraph, MCP, and A2A. A controlled LangGraph experiment
combines an external Reelay temporal monitor with a programmatic report monitor;
ablations show why both are needed for the selected requirements. Differential
checks and cross-adapter tests provide further integration evidence. An LLM-driven
incident-response example illustrates the same enforcement path, while existing
microbenchmarks isolate framework costs. The results support reuse of monitoring
methods at explicit control boundaries, subject to the stated observation and
deployment assumptions.
\end{abstract}
\noindent\textbf{Keywords:} Runtime verification; Agentic AI; Runtime enforcement;
Monitor interoperability.
\section{Introduction}
\label{sec:introduction}

Agentic AI systems combine language models with stateful workflows, external
tools, and other agents. Their behaviour cannot be assessed only from the text
produced by a model. A tool request may change an external system, a state
update may affect later decisions, and a delegated task may continue outside
the local runtime. Relevant observations are spread across model calls,
workflow execution, tools, and communication boundaries such as the Model
Context Protocol (MCP) and Agent2Agent (A2A)~\cite{anthropic2024mcp,linuxfoundation2025a2a}.

Runtime Verification (RV) checks an execution against a
specification~\cite{DBLP:journals/jlp/LeuckerS09}. Earlier work has separated
instrumentation from the monitoring formalism, including
ROSMonitoring~\cite{DBLP:conf/taros/FerrandoC0AFM20} and
RV4Chatbot~\cite{DBLP:journals/corr/abs-2411-14368}. Runtime enforcement for
LLM agents is also already supported by approaches such as
AgentSpec~\cite{AgentSpec2026}. We build on these ideas. Our problem is how to
connect different monitoring methods to the same agent execution while making
their observation, state, and control assumptions explicit.

Consider an incident-response agent that collects diagnostics, creates a
checkpoint, applies remediation, and sends an external report. A temporal
monitor can require approval and a recent checkpoint before remediation. A
separate programmatic monitor can inspect the report destination and content.
They need different inputs and need not use the same verdict vocabulary. The
runtime must combine their decisions without embedding either monitor's
semantics in the workflow. It must also distinguish an attempted remediation
from a completed one: rejecting a request prevents its effect, but the attempt
still belongs to the audit history.

The underlying issues also occur in conventional software. In the agentic
setting, however, a model selects actions and arguments at runtime, a graph
controls local execution, and tools or delegated tasks cross separately
instrumented boundaries. A model's statement that an action is approved is
not approval evidence. Observing a node's output does not by itself provide
control over its propagation. Receiving a remote task update does not imply
control over the remote agent. A reusable interface must expose these
differences rather than give every event the same enforcement meaning.

This paper presents \AgenticRV, a formalism-agnostic architecture for this
interface. Source adapters produce canonical events. Monitor-specific
projections and abstractions produce the input expected by each monitor, and
normalisation maps native verdicts to a common domain. At a boundary that can
retain an operation, a decision manager selects a supported reaction before
the operation is released. A prior vision paper identified RV constraining
LLM-driven systems as one direction in the relation between RV and
LLMs~\cite{DBLP:journals/corr/abs-2511-14435}; here we give a concrete
architecture and examine the conditions under which that direction provides
useful guarantees.

We make three contributions. First, we define an observation and control
contract: we characterise property-sufficient observations, establish a
normalised-response condition for monitor replacement, and give a conditional
prevention argument. Second, we implement the interface for LangGraph, MCP,
and A2A, identifying which boundaries support observation and which support
interposition. We also describe the application-specific configuration work.
Third, we evaluate joint monitoring with an external temporal monitor and a
programmatic monitor, including cross-adapter control, ablations, temporal
differential checks, and protocol-adapter tests. An LLM-driven example illustrates integration, while
microbenchmarks measure the framework costs separately from monitor algorithms.

The contribution is a runtime interface and its contracts, rather than a new
specification language or a verification method for model reasoning. The
guarantees concern visible operations at installed boundaries. They depend on
the observation function, monitor, policy, and faithful adapter enforcement.

\section{Architecture}
\label{sec:architecture}

\begin{figure}[t]
\centering
\begin{tikzpicture}[
  box/.style={draw,rounded corners=1pt,align=center,font=\small,
    text width=2.2cm,minimum height=1.4cm,inner sep=4pt},
  flow/.style={-{Latex[length=1.7mm]},thick},
  control/.style={flow,dashed},
  lab/.style={font=\footnotesize,fill=white,inner sep=1.5pt}]
\node[box] (source) at (0,0) {Agent execution\\{\footnotesize graph, tools, protocols}};
\node[box] (adapters) at (3.0,0) {Source adapters\\{\footnotesize LangGraph / MCP / A2A}};
\node[box,text width=2.6cm] (events) at (6.3,0) {Canonical event layer\\
 {\footnotesize order, subscription,\\projection, abstraction}};
\node[box] (monitors) at (9.6,0) {Monitors\\{\footnotesize temporal / programmatic\\other input domains}};
\node[box,text width=2.6cm] (policy) at (6.3,-2.1) {Decision manager\\
 {\footnotesize normalised verdicts,\\policy, live capabilities}};
\draw[flow] (source)--(adapters);
\draw[flow] (adapters)--(events);
\draw[flow] (events)--(monitors);
\draw[control] (monitors.south)|-node[lab,near end,above]{responses}(policy.east);
\draw[control] (policy.west)-|node[lab,near start,above]{reaction}(adapters.south);
\draw[control] (adapters.south west)--++(0,-0.35)-|
 node[lab,near start,below]{intervention}(source.south);
\end{tikzpicture}
\caption{The event layer is shared; each monitor has its own observation
function. Solid arrows carry observations and dashed arrows carry decisions
and interventions. Only installed interposing boundaries can retain an
operation for a preventive decision.}
\label{fig:architecture}
\end{figure}
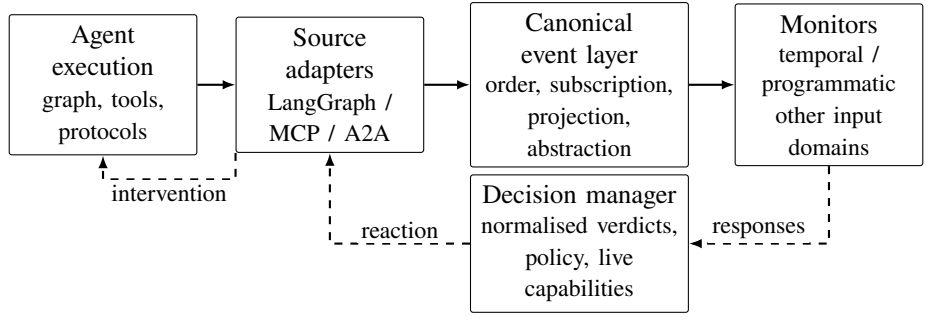

We use \emph{agentic AI system} for an application in which model-driven
components select actions, update state, invoke tools, or delegate tasks.
Figure~\ref{fig:architecture} separates the source adapters, canonical event
layer, monitors, and decision manager. An adapter understands its source but
contains no property-specific logic. A monitor understands its input domain
but need not understand LangGraph, MCP, or A2A.

\paragraph{Instrumentation and events.}
Adapters either observe an execution or interpose on an operation. An
\emph{observation event} reports activity without retaining the operation.
A \emph{decision event} is emitted before a controlled operation and retains
it while synchronous monitors are consulted. Event kind and lifecycle phase
do not alone establish control: an MCP request with phase \tool{before} is
still passive if the adapter is configured for observation only.

Canonicalisation standardises the event envelope, not every payload. The
envelope identifies the source, kind, timestamp, lifecycle phase, and available
execution scope. Its payload records the data exposed by that source. For
example, a summarising node can produce start and completion events containing
its input and returned summary. The next node's input can expose the propagated
value. This does not require a separate event for a value moving along an edge.
Route metadata, when available, supplies a further observation of the selected
branch.

The framework adapter covers local execution. The MCP adapter covers wrapped
tool and resource requests. The A2A adapter covers wrapped discovery, messages,
task operations, status updates, artefacts, and cancellation. Protocol
identifiers can relate these observations to the local task. This correlation
does not reconstruct a global causal order or expose remote internals.

\paragraph{Monitor-facing information.}
Each monitor registration specifies subscriptions, projection, abstraction,
verdict normalisation, synchronous or asynchronous execution, and a decision
deadline. Subscriptions select event kinds. Projection selects or redacts
fields; abstraction maps the resulting view to the monitor input. An LTL
monitor may receive Boolean valuations, while a programmatic oracle receives
structured fields. Either stage may discard an irrelevant event. These choices
are part of the property configuration, since filtering can remove evidence
or change an event-counting clock.

\paragraph{Decisions and interventions.}
The decision manager combines normalised verdicts under a deployment policy.
A \emph{backend-neutral reaction} is a decision such as \allow{} or \block{}
whose meaning does not name a framework API. An \emph{intervention} is its
concrete implementation: blocking a native tool suppresses the callable;
blocking an MCP request suppresses forwarding. The adapter advertises the
reactions supported at each boundary, and records the selected intervention
and outcome as another canonical event.

The abstract interface also admits pausing, replacing, requesting approval,
escalating, aborting, and compensating. An adapter need not support all of them.
In particular, a passive callback cannot retroactively block its originating
operation. Later escalation or compensation has a different guarantee.
Arbitration handles multiple monitors and configured failures, including
timeouts and inconclusive results. Fail-open permits an operation without a
usable mandatory verdict; fail-closed selects a preventive fallback when the
boundary and monitor authorisations support it.

\paragraph{Trust and scope.}
We assume correct adapters, event delivery, policy, and property-specific
monitoring code. Projection limits the observation passed through the
interface; it does not isolate malicious Python code running in the same
process. Guarantees are limited to the instrumented boundaries and their
available interventions. Ordinary node execution, internal state merges,
unwrapped clients, and remote-agent internals do not become controllable simply
because related events are available.

\section{Observation and Control Contracts}
\label{sec:formal_model}

We formalise the interface exposed by the adapters. The model does not describe
an agent's internal reasoning or give semantics to an entire framework.

\subsection{Events, Observations, and Monitors}
\label{sec:formal-events}

\begin{definition}[Canonical event and trace]
A canonical event is
$e=\langle\iota,a,k,t,d\rangle\in
\mathsf{Id}\times\mathcal A\times\mathcal K\times\mathbb T\times\mathcal D$,
where $\iota$ is its identifier, $a$ its source adapter, $k$ its kind, $t$ its
timestamp, and $d$ its data. A canonical trace is a finite sequence
$\tau=e_1\cdots e_n\in\mathcal E^*$ with distinct event identifiers.
\end{definition}

The data component is an open finite map. It can contain lifecycle phase,
execution scope, protocol identifiers, and source payload. We write
$\mathit{id}(e)$ and $\mathit{data}(e)$ for the corresponding components.
Trace order is delivery order, rather than timestamp order. We assume a
centralised event layer that preserves this order for each monitor. Distributed
trace reconstruction and clock synchronisation are outside this model.
The subset $\mathcal E_{\mathsf{dec}}$ contains events for which an adapter
retains an operation before execution; all other events are observations.

For each monitor $m$, let
\[
\pi_m:\mathcal E\to\mathcal E_m\cup\{\eps\},\qquad
\alpha_m:\mathcal E_m\to\Sigma_m\cup\{\eps\}.
\]
Projection selects the authorised view in $\mathcal E_m$; abstraction maps it
to the input domain $\Sigma_m$. Either function can return $\eps$, discarding
the event. Subscription matching is included in $\pi_m$. Extending
$\alpha_m(\eps)=\eps$, we write $\obs_m=\alpha_m\circ\pi_m$ and
\[
\obs_m(e_1\cdots e_n)=
\bigl(\obs_m(e_1)\cdots\obs_m(e_n)\bigr)\downarrow_{\eps},
\]
where $\downarrow_{\eps}$ removes discarded events. For an LTL monitor,
$\Sigma_m=2^{AP_m}$; another monitor can receive structured values or numerical
features. Canonicalisation does not choose between these domains.

\begin{definition}[Monitor]
A monitor is $M_m=\langle Q_m,q_m^0,\Sigma_m,V_m,\delta_m,\lambda_m\rangle$,
with transition and verdict functions
\[
\delta_m:Q_m\times\Sigma_m\to Q_m,
\qquad
\lambda_m:Q_m\to V_m\cup\{\bot\}.
\]
Here $Q_m$ is its state space, $q_m^0$ its initial state, $\Sigma_m$ its
observation alphabet, $V_m$ its native verdict domain, and $\bot$ denotes
absence of a verdict. On observation $\sigma$, the monitor moves to
$q'=\delta_m(q,\sigma)$ and returns $\lambda_m(q')$.
\end{definition}

The transition and verdict functions may be generated from a specification,
implemented directly, or supplied by another analysis method. The interface
does not establish their correctness. Online and offline monitors can process
the same observation sequences; offline verdicts cannot prevent effects in
the recorded execution.

\subsection{What Must an Observation Preserve?}
\label{sec:sufficiency}

Projection and abstraction must preserve the answer to the specified prefix
question. Let $b:\mathcal E^*\to\mathcal W$ give the intended semantic answer
at each prefix. The answer domain may be $\mathcal W=\{0,1\}$ for a Boolean
safety gate, or $\mathcal W=\{\mathsf T,\mathsf F,\mathsf ?\}$ for three-valued
temporal prefix semantics. It need not coincide with the monitor's native
verdict domain $V_m$. Let $H\subseteq\mathcal E^*$ be the admissible prefixes
of the deployment, including its assumptions about event sources and
workflow scope.

\begin{definition}[Property-sufficient observation]
An observation function $\obs_m$ is sufficient for $b$ on $H$ if
\[
\forall\tau,\tau'\in H:\quad
\obs_m(\tau)=\obs_m(\tau')\ \Longrightarrow\ b(\tau)=b(\tau').
\tag{1}\label{eq:sufficient}
\]
\end{definition}

\begin{proposition}[Exact observation-based answers]
There exists a function
\[
\widehat b:\obs_m(H)\to\mathcal W,\qquad
b=\widehat b\circ\obs_m\text{ on }H,
\]
if and only if \eqref{eq:sufficient} holds.
\end{proposition}
\begin{proof}
If $b=\widehat b\circ\obs_m$, equal observations yield equal answers.
Conversely, define $\widehat b(w)=b(\tau)$ for any $\tau\in H$ with
$\obs_m(\tau)=w$. Equation~\eqref{eq:sufficient} makes this choice
independent of the representative.
\end{proof}

The obligation depends on the chosen answer function. If $b$ identifies
bad prefixes, $b(\tau)=0$ means that $\tau$ is not a bad prefix; it does not
assert that the property is satisfied. If $b$ instead gives three-valued
prefix semantics, sufficiency must also preserve the distinction between
satisfaction and inconclusiveness.

This is an information condition, not a monitor-synthesis algorithm.
If two prefixes have the same view but require different semantic answers,
no exact monitor using only that view can distinguish them. For a Boolean
safety gate, rejecting an ambiguous request can avoid permitting an unsafe
candidate, but may also reject a safe one. This conservative choice does
not recover the lost distinction.

For example, if approval is represented by a trusted Boolean, the temporal
monitor need not receive the credential used to establish it. Sufficiency then
depends on the trusted validator, binding approval to the relevant incident,
and preserving the completion event. If the property concerns the report
destination, removing that destination without retaining an adequate
classification makes the observation insufficient. A scope identifier or a
separate monitor instance is also needed when the property distinguishes tasks.
Otherwise approval for one task can become indistinguishable from approval
for another.

Filtering affects temporal meaning as well as data. A bound of six observations
is measured in the monitor's sequence, after discarded events are removed.
Discarding diagnostic observations can make approval appear more recent.
Changing the subscription or abstraction therefore requires checking
\eqref{eq:sufficient} again; it is not merely a performance optimisation.

\subsection{Normalised Verdicts and Reactions}
\label{sec:formal-reactions}

Let $\nu_m:V_m\to\widehat V$ map native verdicts to a common domain, for example
\[
\widehat V=\{\permit,\reject,\mathsf{inconclusive},\mathsf{error}\}.
\]
A response records the event identifier, monitor, native verdict, normalised
verdict, and common control metadata. This metadata includes response status,
mandatory participation, priority, and authorised reactions. Native verdicts
and diagnostics are retained for audit.

Let $R_e$ be the responses collected for event $e$, and let $\eta(R_e)$ remove
native verdicts and diagnostics while retaining normalised verdicts and the
common control metadata. If replacement is enabled, an authorised replacement
proposal is also explicit control data. For adapter $a$, let
$C_a(e)\subseteq\mathcal R$ be the supported reactions. We require the policy
to have the form
\[
P(e,R_e,C_a(e))=\overline P(e,\eta(R_e),C_a(e))\in C_a(e).
\tag{2}\label{eq:policy}
\]
Thus, native values may be stored in a response without being interpreted by
arbitration. A policy that branches on a monitor-specific diagnostic or native
verdict would violate this contract.

\begin{proposition}[Decision preservation under monitor replacement]
For the same decision event, capabilities, and policy, replacing monitors
preserves the selected reaction whenever $\eta(R_e)$ is unchanged.
\end{proposition}
\begin{proof}
Both executions give $\overline P$ the same arguments in
\eqref{eq:policy}, so a deterministic policy returns the same reaction.
\end{proof}

The condition includes failure and deadline status, priority, authorisation,
and any replacement proposal. Equality of successful verdicts alone is
insufficient if a replacement monitor times out. For a fixed event stream,
the result applies at every decision point. It does not require the two
monitors to implement the same formalism, only the same control-facing
responses in that deployment.

\subsection{Attempts, Effects, and Prevention}
\label{sec:formal-interventions}

An attempt and its effect are different observations. A rejected request
belongs to the canonical audit trace but not to the sequence of completed
effects. Suppose a monitor checks
$\mathbf G\neg\mathsf{unapprovedRequest}$ on an attempt
trace and enters an absorbing violation state when an unapproved request is
made. Blocking the request does not erase that attempt, so this monitor
correctly remains violated. Under a policy that blocks every violation, it
will also reject later requests. This is fail-stop behaviour, not evidence
that those later requests are unsafe.

A recoverable gate instead answers whether the \emph{current} request is
permitted from the relevant history. It can reject an unapproved attempt,
retain the evidence that actually occurred, and permit a later valid request.
An audit monitor may still retain the historical violation. Another option is
to scope a fail-stop monitor to a task that is intentionally aborted. Resetting
a monitor after each rejection is not a general repair: it can remove approval,
revocation, or other history still needed for safety.

Let $u\in\mathcal B^*$ be the completed-effect prefix and $x\in\mathcal B$
the pending effect. For an allow/block gate, define the intended query by
$ux\in L$, where $L\subseteq\mathcal B^*$ is a prefix-closed safety language
with $\eps\in L$. A monitor may implement this query from the full audit
history, provided its state represents the relevant completed effects and
current attempt correctly. The interface does not assume transactional
rollback of arbitrary monitor state.
If a requirement depends on time or other observations, these must also be
represented in the safety model; erasing rejected attempts does not by itself
preserve that requirement.

\begin{proposition}[Conditional prevention]
Assume that: (i) every effect passes through an installed decision boundary;
(ii) decisions and effects are serialised, so the gate checks the actual
completed prefix $u$; (iii) every candidate $ux\notin L$ is rejected by a
mandatory synchronous gate; and (iv) rejection or failure blocks release,
with authorised and supported blocking faithfully applied by the adapter.
Then every completed-effect prefix belongs to $L$.
\end{proposition}
\begin{proof}
Initially $u=\eps\in L$. A blocked request leaves $u$ unchanged. If $x$ is
released, condition (iii), together with (iv), excludes $ux\notin L$, so
$ux\in L$. Induction over completed effects proves the claim.
\end{proof}

These assumptions identify the assurance boundary. A subscription miss, an
unsound abstraction, fail-open behaviour, an unsupported preventive reaction,
an unwrapped call, or a failed intervention can invalidate the argument.
Concurrent operations additionally need an atomic check-and-release discipline
or a property-specific concurrency argument; an ordered event log alone does
not establish it. Replacement requires checking the substituted effect, and
is not covered by this allow/block result.

Each adapter implements a reaction through
$\gamma_{a,e}:C_a(e)\to\mathcal I_a$, where $\mathcal I_a$ is its set of
interventions. Prevention requires this intervention to succeed while the
operation is held. Events emitted after dispatch, asynchronous verdicts, and
late responses can support later reactions but cannot prevent that occurrence.

\section{Implementation and Instantiation}
\label{sec:implementation}

\AgenticRV{} is a Python package whose core depends on neither an agent
framework nor a monitoring library. LangGraph, MCP, and A2A are optional
integrations. The inspected implementation is version 0.1.0. We distinguish
implemented operations, executable integration tests, and measured performance
coverage below; support for an operation is not evidence of an experiment
using it.

\subsection{Runtime, Monitor Interface, and Adapter Coverage}

Immutable \tool{CanonicalEvent} objects implement the formal envelope. A private
asynchronous loop assigns an increasing sequence number, orders delivery, and
maintains one serial queue per monitor. Matching precedes field projection and
redaction; user abstraction receives only the projected view. Python callables,
stateful objects, and JSON-over-HTTP monitor services share the same request
interface. Native verdicts are stored alongside their normalised values.
The built-in policies use common verdicts, status, priority, authorisation, and
explicit replacement proposals, rather than the native verdict or diagnostics.

The runtime supports deny-overrides, unanimous, and priority arbitration.
Failures, missing verdicts, and deadlines enter the configured failure policy.
An asynchronous or late contribution cannot change a finalised decision.
A delivery gap marks the affected monitor desynchronised rather than silently
using stale state. Recording and replay support fresh offline analysis with
the same observation functions; replay does not invoke live interventions.

Table~\ref{tab:coverage} identifies the control boundaries. The LangGraph proxy
injects a callback handler into the runnable configuration. Model events are
obtained only from callback-aware LangChain components; a raw provider HTTP
client needs separate instrumentation. Node inputs and outputs expose some
internal state propagation, but the proxy does not guard every reducer or
state merge. Selected routes are observations when callback metadata exposes
them, not independently blockable graph edges.

\begin{table}[t]
\caption{Implemented boundaries. D denotes a held decision; O denotes passive
observation. Decision boundaries support allow, log, block, and replace.
Protocol requests are D in interposing mode and O in passive mode.
Coverage requires use of the wrapper or handler.}
\label{tab:coverage}
\centering\small
\begin{tabularx}{\textwidth}{@{}L{1.75cm}XL{2.0cm}@{}}
\toprule
Adapter & Operation or evidence & Class \\
\midrule
LangGraph & Proxied graph invocation; explicit \tool{update\_state}; guarded
native tool & D before dispatch \\
 & Nodes, subgraphs, exposed routes, callback-aware models and tokens,
retrievers, state reads, checkpoints, interrupts, retries, custom events,
completions and errors & O \\
MCP & Wrapped initialisation, tool call, resource read & D or O \\
 & Negotiated capabilities, results, errors, cancellations & O \\
A2A & Wrapped discovery, outgoing message, task request, subscription,
cancellation request & D or O \\
 & Status, artefact and streaming updates, responses and failures & O \\
Runtime & Applied intervention and its outcome & O \\
\bottomrule
\end{tabularx}
\end{table}

MCP support includes a client-session proxy for initialisation, tool calls, and
resource reads. A2A support includes a client proxy and an SDK interceptor for
outgoing operations and returned task facets. These adapters are implemented,
not future work. The executed tests include a real in-process MCP session and
a real A2A SDK client using a controlled transport
(Section~\ref{sec:validation}). They establish bounded integration behaviour;
they do not establish remote-service throughput or correctness of opaque agents.

\subsection{What Is Manual, and What Is Reused?}
\label{sec:configuration-effort}

Instantiation has three layers. The adapter supplies framework and protocol
event kinds and supported reactions. The monitor integration supplies its
input domain, transition function, and native-verdict normalisation. The
application supplies requirements, trusted evidence, relevant scopes, field
projection, abstraction, and deployment policy. The architecture reuses the
first two layers where possible; it does not automatically derive the third
from natural-language requirements.

For the incident-response application, the existing LangGraph adapter is
installed once and six effect functions are guarded. Application code validates
approval and classifies report contents. The Reelay registration subscribes
to \tool{langgraph.tool}, includes 17 field paths, explicitly excludes four raw
fields, and maps two native verdicts. One abstraction uses the tool name,
lifecycle phase, and five trusted Boolean fields to produce a total valuation.
Approval and checkpoint are true on successful completion; remediation and
reporting are true on held attempts. The supplied abstraction spans 57 source
lines, with a 15-line classification helper; the stateful Reelay facade spans
76 lines, including lifecycle hooks and evidence recording. These are source
inventories including comments and blank lines, not development-time estimates.

The distinction matters in reuse. Adding a property over existing fields
requires a specification or monitor and its registration. Adding a genuinely
new observation may require application instrumentation or an adapter change.
Replacing Reelay requires another facade and normaliser, but not modifications
to the LangGraph wrapper. A content monitor may legitimately need raw report
text that the temporal monitor must not receive. Projection declarations are
therefore reviewed per property rather than copied indiscriminately.

Subscriptions and field projection are declarative; abstraction and trusted
classification may require Python code. Native verdict mappings can be declared
as a table, while numerical verdicts can use configured thresholds. Validation
checks deployment structure and callable interfaces, not semantic sufficiency.
The condition in Section~\ref{sec:sufficiency} supplies the corresponding review
obligation. Time bounds, task scoping, failure policy, and monitor recovery
remain explicit engineering choices.

The case study and controlled experiments use one monitor instance per logical
workflow and serial effect requests. No automatic isolation of arbitrary
interleaved tasks is claimed. A deployment with concurrent tasks must introduce
suitable per-task monitor state and preserve the scope information needed by
its requirements.

\section{Evaluation}
\label{sec:evaluation}

We separate four questions: (Q1) can monitors with different inputs and native
verdicts jointly control the same workflow? (Q2) does the temporal integration
preserve the selected specification, including its boundaries and recovery?
(Q3) which implemented adapter and runtime behaviours have executable
validation? (Q4) what are the costs of instrumentation and dispatch apart from
non-trivial monitor computation? The first three questions use controlled
validation, while the fourth uses microbenchmarks. The validation and timing
sets use different environments and are not pooled.

\subsection{Joint Monitoring and Controlled Ablations}
\label{sec:joint-monitoring}

The controlled experiment uses a real compiled LangGraph, the existing
LangGraph adapter, and two mandatory synchronous monitors. Each operation is
executed through a guarded tool and records its effect in an in-memory ledger.
The driver supplies deterministic action sequences, with one monitor state per
workflow. This isolates monitoring and enforcement from action selection and
does not use an LLM. No changes to the core runtime or adapters were needed.

\paragraph{Temporal monitor.}
The first monitor uses Reelay 25.0.0~\cite{DBLP:journals/lmcs/Ulus26} with the
fixed Rye specification used by the incident-response example. At monitor step
$i$, it checks the current predicate
\[
\chi_i=
\bigl(\mathit{remediate}_i\Rightarrow
 (\mathsf O_{\leq6}\mathit{approval}\land
  \mathsf O_{\leq3}\mathit{checkpoint})_i\bigr)
\land(\mathit{report}_i\Rightarrow\neg\mathit{secret}_i).
\]
The past-time operator $\mathsf O_{\leq k}$ means that its argument occurred
at a distance of at most $k$ monitor steps, including the current step.
One step is produced for each held attempt and each successful tool completion.
Approval and checkpoint are true only at completion; remediation and reporting
are true at their held attempts. Other tool observations still advance this
logical clock. Non-tool callbacks and discarded phases do not.

The bounds 6 and 3 are illustrative validation parameters, not elicited
incident-response requirements. They distinguish the baseline approval distance
of five steps from the stale distance of seven. The boundary experiment below
also tests the exact limits. A deployment must derive suitable bounds from its
requirements and observation clock; physical-time freshness would require
timestamp-based monitoring instead.

Reelay receives a total Boolean valuation and the action name. It does not
receive raw report text or its destination. Its native \tool{satisfied} and
\tool{violation} verdicts map to \permit{} and \reject. The fixed formula
checks $\chi_i$, rather than an absorbing verdict for
$\mathbf G\chi$ over all attempts. Rejection therefore need not prevent a
later request satisfying its current guard. Historical violations remain in
the recorded responses.

\paragraph{Programmatic monitor.}
The second monitor receives only a reporting request's destination and raw
report text. It checks an exact HTTPS hostname allowlist and rejects reports
containing the literal \tool{INTERNAL\_API\_KEY=}. Its native verdicts are
\tool{ok}, \tool{unsafe\_destination}, and \tool{secret\_content}. The latter
two map to \reject. It receives no approval or checkpoint history and cannot
enforce the temporal requirement. The literal is a controlled stand-in for a
content rule, not a general secret detector.

The two monitors therefore differ in their input, state, implementation, and
native vocabulary. Both use the same canonical tool events. The policy blocks
when a mandatory monitor rejects, and both registrations authorise blocking.
Successful completions are delivered in order before the next tool decision.

\paragraph{Paths and ablations.}
We execute seven paths: a safe workflow; missing approval followed by a public
report; stale approval; reporting to an unapproved destination; a secret report;
a secret report whose supplied Boolean classification is false; and a rejected
secret report followed by a public report. The classification mismatch tests
the value of checking raw content with an independently configured monitor:
the temporal monitor's observation is insufficient for this error, whereas
the content monitor detects the chosen literal.

Each path is run with no monitor, Reelay alone, the report monitor alone, and
both. The driver continues after a rejection where the path includes a recovery
request. Table~\ref{tab:ablation} reports 34 tool decisions per configuration,
comprising six unsafe and 28 safe requests under the selected requirements.
The labels are fixed by the controlled inputs and checked against the effect
ledger; they are not obtained from either monitor's verdict.

\begin{table}[t]
\caption{Controlled ablations: seven paths and 34 tool decisions per
configuration. Counts concern the constructed requirements and inputs.}
\label{tab:ablation}
\centering\small
\begin{tabular}{@{}lrr@{}}
\toprule
Active monitors & Unsafe requests executed (of 6) & Safe requests blocked (of 28)\\
\midrule
None & 6 & 0\\
Reelay only & 2 & 0\\
Report monitor only & 2 & 0\\
Both & 0 & 0\\
\bottomrule
\end{tabular}
\end{table}

Reelay alone misses the unapproved destination and the classification mismatch.
The report monitor alone misses missing and stale approval. Together they block
all six unsafe attempts without blocking a safe request in these paths. In the
recovery paths, a public report is permitted after a rejected remediation or
secret report. The rejected effect is absent from the ledger; the subsequent
permitted effect is present. No monitor error, timeout, late response, or dropped
delivery is recorded in the joint runs.

This experiment demonstrates joint integration and a useful separation of
observations. It covers one external temporal monitor and one programmatic
method. It does not establish interoperability with every formalism, quantify
configuration time, or estimate safety rates on unconstrained agent tasks.

\subsection{A Gate Across LangGraph and MCP}
\label{sec:cross-boundary}

A compiled LangGraph invokes local approval and MCP reporting tools. Both
adapters share the runtime and policy. The
official MCP client/server use in-memory streams, with reports recorded in a
server ledger and no core implementation changes.

The additional requirement is approval for the report's own incident:
$\psi_s=\mathit{report}_s\Rightarrow\mathsf O\mathit{approved}_s$.
Approval remains valid for incident $s$ in this experiment, which tests
correlation rather than $\chi_i$'s freshness bounds. The application binds
\tool{scope.trace\_id} before invocation; nested calls inherit it. A facade
maintains Reelay state per incident. Its projection retains kind, phase,
scope, approval completion and MCP tool name. Abstraction turns completed
approval into \tool{approved=true} and a held reporting request into
\tool{report=true}. An approval claim in report arguments supplies no
evidence. The programmatic monitor receives only destination and text,
applying Section~\ref{sec:joint-monitoring}'s content rule.

Seven paths test matching, missing, mismatched and claimed approval,
unapproved destination, secret content, and recovery after rejection.
Each starts with fresh state; scopes may alternate within a path. They contain
eight reporting requests: six unsafe and two safe. Three configurations give
21 executions and 24 reporting decisions (Table~\ref{tab:cross-boundary});
counts exclude approvals and initialisation. The ledger confirms release or
suppression, with no monitor or delivery failures.

\begin{table}[t]
\caption{Cross-boundary validation. Both monitors are active; only temporal
observation or scope partition changes.}
\label{tab:cross-boundary}
\centering\small
\begin{tabular}{@{}lrr@{}}
\toprule
Temporal observation & Unsafe executed (of 6) & Safe blocked (of 2)\\
\midrule
LangGraph and MCP, per incident & 0 & 0\\
Same events, scope erased & 1 & 0\\
MCP events only, per incident & 0 & 2\\
\bottomrule
\end{tabular}
\end{table}

Approval for $A$ followed by reporting for $B$ yields native
\tool{violation}/\tool{ok}, normalised to \reject/\permit; deny-overrides
blocks release. Erasing scope instead permits this report and makes its
history indistinguishable from approval and reporting for $A$: a concrete
counterexample to Equation~\eqref{eq:sufficient}. Observing MCP alone loses
approval evidence and blocks both legitimate reports. With the full view,
a public report is permitted after secret-report rejection without resetting
approval history. Trustworthy scope and approval remain application
obligations; this is controlled integration, not automatic task isolation or
remote-service evaluation.

\subsection{Differential Checks and Adapter Validation}
\label{sec:validation}

We compare the fixed Reelay monitor with an independent Python evaluator of
$\chi_i$. The alphabet contains six symbols: completed approval, completed
checkpoint, a neutral step, remediation attempt, public-report attempt, and
secret-report attempt. We enumerate all 1,554 non-empty words of length at most
four. Comparing every step of every word gives 5,910 verdict comparisons with
no mismatch. This finite enumeration checks the chosen alphabet and short
histories; it is not a proof for unbounded traces.

Short prefixes do not cross every freshness bound. We separately vary approval
age from one to eight and checkpoint age from one to five at a remediation
request. Excluding equal ages, which cannot represent two distinct observations
in this alphabet, gives 35 cases. Reelay permits exactly when the approval age
is at most six and the checkpoint age at most three. The checks include
rejection followed by a permissible observation, confirming the non-absorbing
interpretation of the current guard.

We also change the native-verdict payload while retaining each of the four
common verdicts and the same control metadata. The built-in policy selects the
same reaction in all four pairs, exercising Equation~\eqref{eq:policy}.
This is a regression check of that interface condition, not an independent
proof of the complete implementation.

The selected repository test set contains 179 passing cases, with no skips or
failures. It covers adapters, projection, policy, ordered runtime delivery,
timeouts and late responses, trace handling, and selected executable examples.
The protocol examples use a real in-process MCP client/server session and a
real A2A SDK client with a controlled JSON-RPC transport. The MCP example
executes \tool{echo} but blocks \tool{delete\_records} before the server
function is reached. The A2A example forwards one permitted message and blocks
the rejected message before transport dispatch. Other adapter tests check
passive mode, replacement, lifecycle events, errors, cancellation, and scope.
These are integration checks, not experiments with deployed remote agents.

The controlled validation uses CPython 3.11.16 on Linux, LangGraph 1.2.14,
LangChain Core 1.6.7, MCP 1.30.0, A2A SDK 1.2.2, and Reelay 25.0.0. The
supplement records the dependency snapshot, scripts, specification checksum,
test report, and per-request outcomes. These versions are separate from the
performance environment below.

\subsection{LLM-Driven Incident-Response Example}
\label{sec:qualitative-evaluation}

The LLM-driven integration example connects Ollama to an iterative LangGraph
agent and the same Reelay monitor. The application alternates an agent node
with guarded effect execution. A JSON schema constrains model output to an
enabled action and an untrusted note. The workflow excludes transitions that
are not meaningful in the current state, but the monitor decides whether an
enabled effect satisfies its temporal and reporting guard.
The six guarded operations are triage, approval validation, diagnostic
collection, checkpoint creation, destructive remediation, and external
reporting. Approval validation and content classification are application
functions. The model does not supply the trusted safety propositions. Raw
credentials, reports, incident identifiers, and model notes are excluded from
the temporal monitor's view. Effects are recorded in an in-memory ledger.
\begin{figure}[t]
\centering
\begin{tikzpicture}[x=0.69cm,y=0.70cm,
  marker/.style={circle,draw,fill=white,minimum size=4.6mm,inner sep=0pt,font=\footnotesize\bfseries},
  blocked/.style={marker,fill=black,text=white},
  labeltext/.style={font=\footnotesize}]
\foreach \y/\a/\c/\r in {3/3/7/8,2/-4/5/6,1/3/9/10,0/3/7/8}{
  \pgfmathsetmacro{\lefta}{max(0,\r-6)}
  \pgfmathsetmacro{\leftc}{max(0,\r-3)}
  \fill[black!6] (\lefta,\y-0.33) rectangle (\r,\y+0.33);
  \fill[black!18] (\leftc,\y-0.33) rectangle (\r,\y+0.33);
}
\draw[-{Latex[length=1.7mm]}] (0,4.0)--(11.5,4.0);
\foreach \s in {0,...,11}{\draw (\s,3.93)--(\s,4.07);\node[labeltext,above] at(\s,4.09){\s};}
\node[labeltext,above] at(5.5,4.6){Logical monitor step};
\foreach \y/\last/\label in {3/11/Safe,2/6/Missing approval,1/10/Stale approval,0/10/Secret report}{
  \draw(0,\y)--(\last,\y);
  \node[labeltext,anchor=east] at(-0.35,\y){\label};
}
\foreach \x/\y/\lab in {3/3/A,7/3/C,8/3/R,10/3/P,5/2/C,3/1/A,9/1/C,3/0/A,7/0/C,8/0/R}
  \node[marker] at(\x,\y){\lab};
\foreach \x/\y/\lab in {6/2/R,10/1/R,10/0/P}
  \node[blocked] at(\x,\y){\lab};
\node[labeltext,anchor=west] at(-2.5,-0.9){A: approval complete\quad C: checkpoint complete\quad R: remediation\quad P: report};
\node[labeltext,anchor=west] at(-2.5,-1.55){Light/dark shading: approval/checkpoint windows. Filled requests are blocked.};
\end{tikzpicture}
\caption{Relevant events from the four Reelay--Ollama witnesses.
Unmarked attempts and completions also advance logical time. Shaded regions
show the histories considered at remediation. A blocked request produces no
completion or ledger entry for its effect.}
\label{fig:qualitative-reelay}
\end{figure}
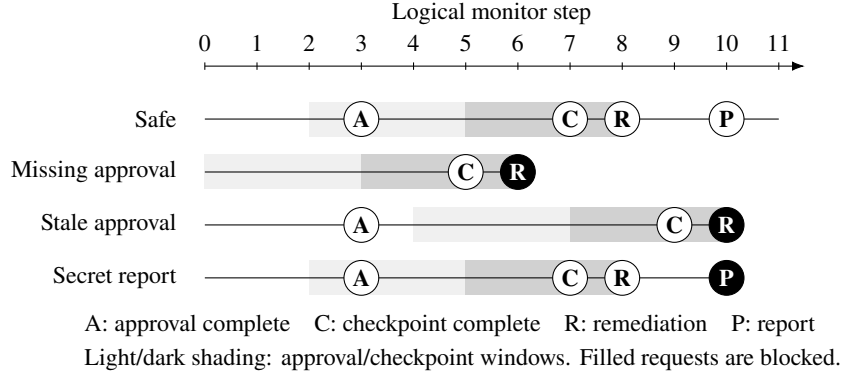

The four executions use \tool{llama3.2:3b} through Ollama 0.12.9,
temperature zero, and fresh graph and monitor state per scenario.
Figure~\ref{fig:qualitative-reelay} shows the relevant events. The safe path
completes remediation and reporting. Missing approval and stale approval each
block remediation. A secret-bearing report is blocked after permitted
remediation, so remediation remains in the ledger while publication is absent.
Each run reaches its intended witness and records no monitor failure or timeout.
The case study's Ollama calls use the raw local HTTP client, so this example
does not generate callback-based \tool{langgraph.model} events.

Scripted providers reproduce these four paths with the same formal histories,
verdicts, and ledgers; the selected tests rerun that deterministic integration.
The four model runs are single witnesses. They
illustrate the execution path, not the frequency of unsafe model choices or
the effectiveness of enforcement over a benchmark of realistic incidents.
The ablations supply the stronger, reproducible evidence about joint
monitoring; they do not turn these four witnesses into a statistical LLM study.

\subsection{Framework Costs}
\label{sec:quantitative-evaluation}

The microbenchmarks isolate instrumentation, dispatch, projection, auditing,
and intervention. Their monitors return immediately in the same process.
They measure framework overhead, not the cost or diversity of substantive
monitor algorithms. Reelay is not part of this timing campaign.

The timing measurements use an Apple M3 Pro with 18\,GiB of memory, Python
3.14.6, \AgenticRV{} 0.1.0, and LangGraph 1.2.9. Each condition runs in
20 fresh processes with three warm-up and five measured batches of approximately
200\,ms. Settled time includes work queued by an operation and drained within
its batch, excluding construction, warm-up, validation, and shutdown. Batches
are averaged within each process before averaging across processes. The 95\%
intervals use 10,000 bootstrap resamples of processes.

\begin{table}[t]
  \centering
  \scriptsize
  \setlength{\tabcolsep}{3pt}
  \caption{Settled wall-clock time per operation. Intervals are 95\%
  worker-bootstrap confidence intervals. Differences are relative to the
  baseline identified for each group. A negative value indicates that less
  work is performed than in the baseline.}
  \label{tab:performance}
  \begin{tabular}{@{}lrrrr@{}}
    \toprule
    Condition &
    Mean ($\mu$s) &
    95\% CI ($\mu$s) &
    $\Delta$ time ($\mu$s) &
    $\Delta$ (\%) \\
    \midrule

    \multicolumn{5}{@{}l}{
      \emph{LangGraph invocation} (baseline: Native)}\\
    Native
      & 222.905  & [221.672, 224.263]   & --        & -- \\
    Proxy--0
      & 919.013  & [901.148, 936.825]   & +696.108  & +312.3 \\
    Proxy--1
      & 1332.660 & [1263.414, 1454.243] & +1109.755 & +497.9 \\

    \addlinespace
    \multicolumn{5}{@{}l}{
      \emph{Guarded tool} (baseline: 0 monitors)}\\
    0 monitors
      & 297.919  & [293.136, 303.084]   & --       & -- \\
    1 monitor
      & 573.731  & [523.640, 666.920]   & +275.812 & +92.6 \\
    2 monitors
      & 628.819  & [618.723, 640.138]   & +330.900 & +111.1 \\
    4 monitors
      & 804.655  & [795.937, 815.548]   & +506.736 & +170.1 \\
    8 monitors
      & 1166.758 & [1157.101, 1176.901] & +868.839 & +291.6 \\

    \addlinespace
    \multicolumn{5}{@{}l}{
      \emph{Projection, 256\,KiB payload}
      (baseline: subscription miss)}\\
    Subscription miss
      & 93.615   & [89.327, 100.752]    & --       & -- \\
    Header
      & 227.628  & [224.821, 230.746]   & +134.013 & +143.2 \\
    Full
      & 339.691  & [337.524, 341.914]   & +246.076 & +262.9 \\
    SHA-256 hash
      & 844.623  & [821.319, 883.086]   & +751.008 & +802.2 \\

    \addlinespace
    \multicolumn{5}{@{}l}{
      \emph{JSONL audit} (baseline: Disabled)}\\
    Disabled
      & 226.770  & [223.583, 229.910]   & --      & -- \\
    Buffered
      & 303.074  & [300.215, 306.146]   & +76.304 & +33.6 \\
    Flush each record
      & 312.647  & [309.175, 316.183]   & +85.877 & +37.9 \\

    \addlinespace
    \multicolumn{5}{@{}l}{
      \emph{Intervention} (baseline: Allow)}\\
    Allow
      & 528.222  & [521.695, 535.113]   & --       & -- \\
    Block
      & 384.008  & [379.657, 388.286]   & -144.214 & -27.3 \\

    \bottomrule
  \end{tabular}
\end{table}

The five groups in Table~\ref{tab:performance} are independent experiments,
not stages of one pipeline. Instrumenting the one-node graph adds
696.108\,$\mu$s over its 222.905\,$\mu$s native execution. A first monitor
adds a further 413.647\,$\mu$s; the complete invocation is 1.333\,ms.
At a guarded tool boundary, increasing from zero to eight immediate monitors
adds 868.839\,$\mu$s. The largest fixed dispatch increment occurs with the
first monitor, followed by roughly linear growth in this setting.

For a 256\,KiB payload, header and full projection add 134.013 and
246.076\,$\mu$s relative to a subscription miss; hashing adds
751.008\,$\mu$s because it reads the complete payload. Narrow views reduce
both exposure and processing work, provided they preserve the property's
evidence. JSONL audit recording adds 76--86\,$\mu$s. Flushing each record
adds 9.573\,$\mu$s over buffering, without an \tool{fsync}.

Blocking is 144.214\,$\mu$s faster than allowing because the blocked path
omits the callable and its completion event. This is a difference in work,
not a performance advantage of rejection. Blocked side effects are absent in
the measured runs, with no recorded monitor errors, timeouts, late responses,
or dropped deliveries.

\paragraph{Limits of the evidence.}
The benchmarks isolate fixed costs, so percentage overheads do not generalise
to model- or network-dominated workflows. They exclude cold starts,
concurrency, remote monitors, MCP/A2A latency and substantive monitor
computation. In-memory validation adds monitor and protocol checks but does
not close these performance gaps. Deployment still requires suitable
approval, content, scope and observation assumptions. No guardrail comparison
is established.

\section{Related Work}
\label{sec:related-work}

Runtime verification separates the observation of an execution from the
analysis of its trace~\cite{DBLP:journals/jlp/LeuckerS09}. Runtime enforcement
additionally controls whether an operation is released. Surveys of
correct-by-construction enforcement in AI distinguish the guarantees of such
mechanisms from checks of an AI component's output
alone~\cite{DBLP:conf/birthday/KonighoferBEP22}. Our prevention argument is
conditional on the installed boundary and monitor contract; it does not
construct a shield from a model of the environment.

\paragraph{Monitoring infrastructures.}
ROSMonitoring separates ROS instrumentation from monitoring backends and does
not prescribe a specification formalism~\cite{DBLP:conf/taros/FerrandoC0AFM20}.
ROSMonitoring 2.0 extends that setting to services and ordered
topics~\cite{DBLP:journals/corr/abs-2411-14367}. RV4Chatbot follows the same
design direction for chatbot interaction protocols, instantiated for Rasa and
Dialogflow~\cite{DBLP:journals/corr/abs-2411-14368}. These are direct
architectural precedents. Formalism agnosticism and adapter-based observation
are therefore not new contributions of \AgenticRV.

We extend this separation to a shared interface for local graph events and
tool/delegation boundaries, with per-monitor observation and normalisation,
explicit boundary capabilities, and a distinction between attempted actions
and effects. The observation-sufficiency condition states which property
information a view must retain. The replacement and prevention propositions
state the limits of the resulting control contract. These mechanisms are
applicable beyond agentic AI; the agentic contribution is their instantiation
across the identified sources and trust boundaries, together with executable
joint-monitor evidence.

\paragraph{Agent guardrails and enforcement.}
NeMo Guardrails provides programmable rails for LLM
applications~\cite{DBLP:conf/emnlp/RebedeaDSPC23}. AgentSpec provides a DSL
combining triggers, predicates, and enforcement actions, and evaluates runtime
constraints on several kinds of LLM agents~\cite{AgentSpec2026}. It already
supports modular runtime enforcement and adaptation to agent frameworks. The
distinction is not that \AgenticRV{} introduces enforcement or reusable
instrumentation, but that its core interface leaves the monitor input, state,
formalism, and native verdict domain to separately registered implementations.

ProbGuard learns a probabilistic execution model and predicts safety risks
to trigger proactive intervention~\cite{ProbGuard2026}; RvLLM uses its own
formal-specification approach to constrain LLM behaviour
\cite{DBLP:conf/nips/ZhangEED25}. These methods have their own guarantees and
evaluation scopes. They are possible sources of monitoring components for an
interoperability layer, subject to appropriate observation and response
bindings. We have not integrated them, and the two-monitor experiment does
not establish compatibility with their implementations.

\paragraph{Specifications and monitor generation.}
Operational and declarative monitoring approaches can be combined for
data-rich traces, as in TP-DejaVu~\cite{DBLP:conf/vmcai/HavelundKOPT24}.
NL2TL translates natural-language requirements into temporal
logic~\cite{DBLP:conf/emnlp/ChenGZF23}. Such work can help obtain a monitor
or specification, while \AgenticRV{} addresses its connection to execution
boundaries. It does not remove the need to establish trusted propositions,
review information loss, or select a failure policy.

AgentHarm evaluates harmful multi-step agent
behaviour~\cite{DBLP:journals/corr/abs-2410-09024}. It motivates evaluation
over full executions, but our controlled paths are not a substitute for such
benchmarks. The present results concern interface integration and conditional
enforcement, rather than comparative end-to-end agent safety.

\section{Conclusion}
\label{sec:conclusions}

\AgenticRV{} connects agent execution boundaries to monitors with different
input and verdict domains. Its contract separates sufficient observation,
normalised arbitration, and supported intervention, while distinguishing a
rejected attempt from a completed effect. LangGraph, MCP, and A2A adapters
implement the interface at explicit boundaries. Controlled ablations show
joint enforcement by Reelay and a programmatic report monitor, including
permitted recovery after rejection. A LangGraph--MCP workflow also tests
approval correlation across incident scopes. Differential checks and adapter tests add
reproducible integration evidence; the microbenchmarks isolate framework costs.
The next evaluation steps are broader agent workloads, monitor configuration
effort, substantive remote monitors, and concurrent check-and-release behaviour.
Distributed trace reconstruction and stronger event-integrity guarantees also
require work beyond the centralised model studied here.

\paragraph{AI usage.}
OpenAI's ChatGPT and Codex assisted with manuscript preparation, diagrams and
validation code; the author remains responsible for verifying their outputs
and for the final content.
\paragraph{Data availability.}
Code:
\url{https://github.com/AngeloFerrando/agentic-rv}.
The supplement contains scripts, a dependency snapshot and results for
commit \texttt{17e0c8c}. Measurement environments are given in
Section~\ref{sec:evaluation}.
\bibliographystyle{plainnat}
\bibliography{main}
\end{document}